\documentclass[final,11pt,authoryear]{elsarticle}
\biboptions{sort}
\makeatletter
\def\ps@pprintTitle{%
 \let\@oddhead\@empty
 \let\@evenhead\@empty
 \def\@oddfoot{\centerline{\thepage}}%
 \let\@evenfoot\@oddfoot}
\makeatother

\usepackage{expex}
\lingset{aboveexskip=2ex,belowexskip=2ex}
\lingset{everyex=\footnotesize}
\usepackage[breaklinks=true]{hyperref}
\usepackage{etoolbox}
\appto\UrlBreaks{\do\-}
\usepackage{xcolor}
\hypersetup{
    colorlinks,
    linkcolor={red!50!black},
    citecolor={blue!50!black},
    urlcolor={blue!80!black}
}
\usepackage{booktabs}
\usepackage{multirow}
\usepackage{array}
\usepackage{ragged2e}

\newcommand{\exs}{\save@counters\refstepcounter{mysub}\renewcommand{\thexnumi}{\arabic{xnumi}\alph{mysub}}\@ifnextchar [{\@ex}{\item}\reset@counters}
\usepackage{tabularx,booktabs}  
\usepackage[margin=2.5cm]{geometry}

\usepackage{amssymb}
\usepackage{breqn}
\usepackage{gb4e}
\usepackage{tabularx}
\usepackage{xspace}
\usepackage{tablefootnote}
\usepackage{subfig}
\noautomath
\PassOptionsToPackage{demo}{graphicx}
\usepackage{subfig}

\let\oldFootnote\footnote
\newcommand\nextToken\relax

\renewcommand\footnote[1]{%
    \oldFootnote{#1}\futurelet\nextToken\isFootnote}

\newcommand\isFootnote{%
    \ifx\footnote\nextToken\textsuperscript{,}\fi}

\newcommand{\CTC}{C3\@\xspace} 

\newcommand{\smCTC}{SOCC-a\@\xspace} 

\newcommand{\NYT}{NYT\@\xspace} 
\newcommand{\YNACC}{YNACC*\@\xspace} 

\newcommand{\xuparrow}[1]{%
  {\left\uparrow\vbox to #1{}\right.\kern-\nulldelimiterspace}
}

\begin{document}

\begin{frontmatter}

\title{Classifying Constructive Comments{\let\thefootnote\relax\footnote{Paper accepted for publication in \href{https://firstmonday.org}{\textit{First Monday}}. Draft version: August 4, 2020.}}}

\author[1]{Varada Kolhatkar\corref{first}}
\cortext[first]{Co-first author}

\author[2]{Nithum Thain\corref{first}}

\author[2]{Jeffrey Sorensen}

\author[2]{Lucas Dixon}

\author[3]{Maite Taboada\corref{cor}}
\cortext[cor]{Corresponding author}

\address[1]{University of British Columbia}
\address[2]{Jigsaw}
\address[3]{Simon Fraser University}

\begin{abstract}

We introduce the Constructive Comments Corpus (\CTC), comprised of 12,000 annotated 
news comments, intended to help build new tools for online communities to improve the quality of their discussions.
We define constructive comments as high-quality comments that make a contribution to the conversation.
We explain the crowd worker annotation scheme and define a taxonomy of sub-characteristics of constructiveness. 
The quality of the annotation scheme and the resulting dataset is evaluated using measurements of inter-annotator agreement, expert assessment of a sample, and by the constructiveness sub-characteristics, which we show provide a proxy for the general constructiveness concept. 
We provide models for constructiveness trained on \CTC using both feature-based and a variety of deep learning approaches and demonstrate, through domain adaptation experiments, that these models capture general rather than topic- or domain-specific characteristics of constructiveness. 
We also examine the role that length plays in our models, as comment
length could be easily gamed if models depend heavily upon this feature. 
By examining the errors made by each model and their distribution by length, we show that the best performing models are effective independently of comment length.
The constructiveness corpus and our experiments pave the way for a moderation tool focused on promoting comments that make a meaningful contribution, rather than only filtering out undesirable content.

\end{abstract}

\begin{keyword}
Content moderation \sep online comments \sep toxicity \sep constructiveness \sep annotation \sep data creation \sep machine learning \sep deep learning
\end{keyword}

\end{frontmatter}

\section{Introduction: Content moderation and constructiveness}
\label{sec:introduction}
One of the key challenges facing online communities, from social networks to the comment sections of news sites, is low-quality discussions.
During the print era, with space carrying high cost, publications used strong
editorial control in deciding which letters from readers should be published.
As news moved online, comments sections were often open and unmoderated,
and sometimes outsourced to other companies. 
One notable exception is \textit{The New York Times}
which has, since 2007,
employed a staff of full-time moderators to review all comments submitted to their website \citep{etim-2017}.
Exemplary comments representing a range of views are highlighted and tagged as \textit{NYT Picks}.
The importance of moderators was unanticipated by many
publishers, with many removing comments sections because they could not adequately moderate the comments.
With vast numbers of online comments, and growing challenges of how social networks manage toxic language, the role of moderators is becoming much more demanding \citep{Gillespie18-COT,Roberts19-BTS}.
There is thus growing interest in developing automation to help filter and organize online comments for both moderators and readers \citep{park_supporting_2016}.

Comment moderation is often a task of filtering out, i.e., deleting toxic and abusive comments, the `nasty' part of the Internet \citep{Chen17-OIA}. Research presented at the Abusive Language Workshops\footnote{\url{https://sites.google.com/view/alw3}} often explores methods to automate the task of detecting and filtering abusive and toxic comments, what \citet{Seering19-DUI} define as \textit{reactive} interventions. 
We propose the flipside of that task, the promotion of constructive comments,
a form of \textit{proactive} intervention \citep[see also][]{jurgens-etal-2019-just}. While filtering will always be necessary, we like to think that what we define as constructive comments are promoted and highlighted, a positive contagion effect will emerge \citep{West14-NNP,Meltzer15-JCA}. 
There is, in fact, evidence that nudges and interventions have an impact on the civility of online conversations and that a critical mass effect takes place with enough polite contributors. \citet{Stroud11-NNT} showed that a `respect' button (instead of `like' and `dislike') encouraged commenters to engage with political views they disagreed with. Experiments indicate that having more polite posts highlighted leads to an increased perception of civility \citep{Grevet16-BNO} and that commenters exposed to thoughtful comments produce, in turn, higher-quality thoughtful comments \citep{Sukumaran11-NIO}. 
Evolutionary game models also support the hypothesis that a critical mass of civil users results in the spread of politeness in online interactions \citep{Antoci16-CVI}.

\citet{Shanahan18-JOC} argues that news organizations ought to be engaged in collecting and amplifying news comments, including seeking out diverse participants and including varied perspectives. The identification of constructive comments can become another tool to help news outlets in fostering better conversations online. The dataset and experiments we present in this paper contribute to that effort. 

To illustrate the various choices one can make when considering the quality of a comment, we show a potential spectrum of comments and their quality in Table \ref{tab:constructiveness_spectrum}. 
At the bottom of the table, we see both negative and positive comments that are non-constructive, because they do not seem to contribute to the conversation. 
The middle comment is not necessarily constructive; it provides an only opinion, and no rationale for that opinion. Such non-constructive, but also not toxic comments, are little more than backchannels and do not contribute much to the conversation \citep{Gautam19-HTC}.

The comment about Trump and Clinton may be perceived as constructive, but it contains abusive language that makes it harder to embrace. Finally, we view the top comment as constructive, because it presents a reasoned opinion, supported by personal experience.

\begin{table*}[htb]
{\small
$\rotatebox[origin=c]{90}{constructiveness}%
\xuparrow{3.8cm}
\hspace{0.4cm}
\centering
    \begin{tabular}{p{2cm} p{12.5cm}}
     \toprule
     Label  &   Example\\
     \midrule
     Constructive &  Simpson is right: it's a political winner and a policy dud - just political smoke and mirrors. Mulcair wants Canada to adopt a national childcare model so he can hang on to seats in Quebec, that's all. Years ago I worked with a political strategist working to get a Liberal candidate elected in Conservative Calgary. He actually told his client to talk about national daycare - this was in the early 90's. The Liberal candidate said, 'Canada can't afford that!' to which the strategist responded 'Just say the words, you don't have to actually do it. It'll be good for votes.' I could barely believe the cynicism, but over the years I've come to realize that's what it is: vote getting and power politics. Same thing here.
     {\scriptsize\url{http://www.theglobeandmail.com/opinion/daycare-picks-up-the-ndp/article21094039/}}\\
     \midrule
     \vtop{\hbox{\strut Constructive}\hbox{\strut (toxic)}}
     &   Please stop whining. Trump is a misogynist, racist buffoon and perhaps worse. Clinton is, to put it in the most polite terms possible, ethically challenged and craven in what she will tolerate in her lust for power. Neither of them is a stellar representative of their gender. Next time, put up a female candidate who outshines the male, not one who has sunk to his same level. Simple.
     {\scriptsize\url{https://www.theglobeandmail.com/opinion/thank-you-hillary-women-now-know-retreat-is-not-an-option/article32803341/}}\\   
     \midrule
     \vtop{\hbox{\strut Opinion}\hbox{\strut (no justification)}}
     &   Please do not print anything Dalton writes and do not report anything he says or does until he does his time in prison.
     {\scriptsize\url{http://www.theglobeandmail.com/opinion/being-clean-and-green-comes-with-a-cost/article27730073/}}\\
     \midrule
     \vtop{\hbox{\strut Non-constructive}\hbox{\strut (positive)}} & Another wonderful read! Thanks Maggie!
     {\scriptsize\url{http://www.theglobeandmail.com/opinion/david-gilmour-an-agent-of-the-patriarchy-oh-please/article14570359/}}\\           
     \midrule
     \vtop{\hbox{\strut Non-constructive}\hbox{\strut (insulting)}}      & Another load of tosh from a GTA Liberal.
     {\scriptsize\url{https://www.theglobeandmail.com/opinion/thank-you-hillary-women-now-know-retreat-is-not-an-option/article32803341/}}\\
     \bottomrule \\
    \end{tabular}
$
\caption{Constructiveness spectrum. The first row represents the most constructive and the last row represent the most non-constructive ends of the spectrum. Links are to the article that triggered the comment.}
\label{tab:constructiveness_spectrum}
}
\end{table*}

Existing approaches to classifying the quality of online content primarily focus on various forms of toxicity in comments \cite[e.g.,][]{Kwok:2013,Waseem16-HSO,Nobata16-ALD,Davidson17-AHS,wulczyn:2016,Pushkar19-ALD}.
Some research has examined the characteristics of constructive comment threads \citep{Napoles17-FGC} and our previous work \citep{Kolhatkar17a} suggests that we need to consider constructiveness along with toxicity when moderating news comments, because some toxic comments may still be constructive, as shown in the second row of Table \ref{tab:constructiveness_spectrum}.

In this paper, we first introduce and evaluate a new annotation scheme for crowd workers to rate the \textit{constructiveness} of an individual comment. 
This is intended to capture readers' ability to assess comments that add value to the article being commented on.
We also create a taxonomy of sub-characteristics, attributes related to constructiveness to further evaluate our definition. 
We then annotate our corpus with both binary constructiveness labels and sub-characteristics through crowdsourcing.  
Additionally, we record when a crowd worker indicates agreement with the view being expressed in the comment, to test whether crowd workers assign constructiveness more often to comments they agree with.

The annotated corpus constitutes what we term the \textit{Constructive Comments Corpus (\CTC)}, which consists of 12,000 news comments in English enriched with a crowd-annotated fine-grained taxonomy of constructiveness and toxicity scores.\footnote{In line with \cite{bender-friedman-2018-data} the discussion in Section \ref{sec:data_and_annotation} is aimed at addressing the key points of a data statement. It is incomplete as we do not have access to information about specific speaker or annotator demographics, though trends about the latter can be found in analyses of crowdworker demographics \cite{posch2018characterizing}.}
This is the largest corpus of comment constructiveness annotations that we know of, being approximately 10 times larger than the one described in \citet{Kolhatkar17a}.

To illustrate the use of this dataset, we develop both classical feature-based and deep learning systems to classify constructive comments. 
Our methods make the classification decision based solely on the text of the comment, without relying on the commenters' past behaviour. 
This is useful when such information is limited as well as when one wishes to evaluate a comment on its own merit alone. 
We also show the transferability of such models across domains, training on one dataset and evaluating on another.

We outline some challenges and new results in modelling constructive comments. 
Our analysis highlights that naive models of constructive comments will have length as an overwhelmingly important feature.
This obviously produces a trivially tricked model which is likely to be of limited practical value. 
By exploring several contemporary deep learning architectures, we show that some architectures are robust to this effect, such as Convolutional Neural Networks (CNNs) or Transformer-based models, which represent entire comments in ways that do not depend directly on the input length.

Finally, we note that all code and data are released under open-source and public domain licenses. See Section \ref{sec:conclusion} for links.

\section{Related work: Identifying high-quality comments}
\label{sec:related}
Much of the focus on online comments is on their negative characteristics, including toxicity, abusive language, hate speech and polarization \citep{Warner12-DHS,wulczyn:2016,Saleem16-AWO}. We take an interest in the positive aspects of online comments, those that make a comment worthwhile and help promote engagement. 
High-quality comments on an online publication foster a sense of community, in particular if comments are perceived to be moderated \citep{Meyer14-IME}. 

Research into what constitutes a high-quality comment has shown that they tend to be constructive, i.e., they contribute an opinion or point of view, and provide reasons or background for that opinion. A 2015 study of the New York Times Picks \citep{Diakopoulos15-PTN}
showed that Pick comments, compared to non-Picks, have
higher argument quality, are more critical, show internal coherence (i.e., they do not excessively rely on context), share some personal experience, are thoughtful, and are readable. 
Automatically computed measures of readability indicate that Picks require a higher reading level. 
Interestingly, length was a significant factor in this study, with NYT Picks having an average of 127.2 words per comment (as opposed to 81.7 words for non-Picks). Additional considerations in the commercial world of online publishing likely exist, including increasing reader engagement.


We define high-quality comments as comments that are \textit{constructive}. Constructiveness is a subjective term, infused with moral, historical, and political biases. 
Previous work that has tackled the issue of constructiveness in online comments and discussions offers a range of definitions. \citet{Niculae16-CMO} define a constructive online discussion as one where the team involved in the discussion improves the potential of the individuals. That is, the individuals are better off (in a game) when their scores are higher than those they started out with. The definition of \citet{Napoles17-FGC} is characterized as more traditional: comments that intend to be useful or helpful. They define constructiveness of online discussion in terms of ERICs---Engaging, Respectful, and/or Informative Conversations. In their annotation experiment, those were positively correlated with informative and persuasive comments, and negatively correlated with negative and mean comments. 
\citet{Loosen18-MSO} hear from journalists and content moderators that comments to be promoted are those that present new questions, arguments or viewpoints. There may be, however, some disagreement in which comments are rated as constructive by moderators as opposed to readers \citep{Miro20-TCG}. 
Our definition is based on the existing literature and on our previous work. In \citet{Kolhatkar17a,Kolhatkar17b}, we surveyed online users and built a definition based on their answers. Online users characterize constructive comments as posts that intend to create a civil dialogue through remarks that are relevant to the article and not intended to merely provoke an emotional response. Furthermore, constructive comments are typically targeted to specific points and supported by appropriate evidence. This definition applies narrowly to online news comments, which is why a contribution is seen as providing an opinion with some justification or evidence for the opinion. It also has many points in common with the definitions in \citet{Berry17-DQD}, based on rater's intuitions about Facebook comment quality.

Given the large volume of comments that a publisher may want to moderate and label so that they can promote constructive comments, the task is a natural candidate for automatic content moderation. Indeed, most of the work on comment identification and moderation tends to take a text classification approach, modelling characteristics of good and bad comments or threads using supervised techniques. We focus here on the task of moderating individual comments using information in the comment itself (i.e., not metadata about the comment or the author). Thus, we rely on textual aspects within the comment, but additional measures of constructiveness are possible, such as degree of connection between comment and article, as a proxy for relevance. See, for instance, research on probabilistic topic models \citep{Hoque19-ITH} or on word overlap in article and comment \citep{Risch18-DON}.

While most of the existing work in automated comment moderation focuses on filtering, i.e., blocking or deleting `bad' comments, some of the techniques can be applied in the complementary task of finding and promoting `good' comments. Classic approaches include feature-based classifiers, typically using Support Vector Machines \citep{Nobata16-ALD,Davidson17-AHS} or logistic regression \citep{Risch18-DON,Waseem16-HSO} with features such as character and word n-grams, average word length of words in a comment, length of comment or linguistic features (modals and hedges, dependency parses). 

Word embeddings are popular in the toxicity detection literature, with methods ranging from averaging pre-trained word embeddings \citep{Nobata16-ALD,Orasan18-ALI} to more contextual models using embeddings from paragraph2vec \citep{Djuric15-HSD}. 
Deep learning approaches employ recurrent neural networks \citep{Pavlopoulos17-DLF,Pavlopoulos17-DAT} or various forms of convolutional neural networks \citep{Zhang18-DHS,Gambaeck17-UCN}. See also \citet{Schmidt17-ASO} for a general survey. 

Although many of these approaches are applicable in the complementary task of identifying constructive comments, a slightly different approach is needed. The task is not just one of pinpointing abusive words and expressions, no matter how subtle; it is about detecting that an argument is being made, that evidence is being provided, and that the comment contributes to the conversation. 
This is also why toxicity detection methods that use sentiment analysis or polarity of the words in the comment are not useful in this context \citep{Orasan18-ALI,Sood12-AIO}.

More specifically with the goal of identifying constructive comments, \citet{Napoles17-FGC} use annotated threads (as opposed to individual comments, as we do here) and model constructiveness using different machine learning models, including a linear model with various features (averaged word embeddings, counts of named entities or length) and a neural model (CNN). Their best performance is with a feature-based model, with an F1 score of 0.73. \citet{park_supporting_2016} aim to distinguish NYT Picks from non-picks using a Support Vector Machine classifier with features that are a mix of comment-based (relevance to the article via word similarity, length, readabiliy) and user-based (number of comments the user has posted, average length of those, history of recommendation by others). Cross-validation precision for this system was 0.13, with 0.60 recall, both relatively low because of a small dataset.

In this paper, we add to this literature by introducing a new dataset for constructiveness, which served as the basis for several experiments to build an automatic classifier for comments. Our experiments show that this is a rich and very useful dataset, the largest to date with such annotations, and that the task of identifying high-quality comments poses interesting challenges for the research community.

\section{Data and annotation}
\label{sec:data_and_annotation}
We present \CTC (Constructive Comments Corpus), a corpus of 12,000 online news comments enriched with constructiveness annotations and toxicity scores.
The 12,000 comments were drawn from the SFU Opinion and Comments Corpus (SOCC), a freely available resource,\footnote{\url{https://github.com/sfu-discourse-lab/SOCC}} which contains a collection of opinion articles and the comments posted in response to the articles \citep{Kolhatkar-SOCC-2019}. The articles include all the opinion pieces published in the Canadian English-language newspaper \textit{The Globe and Mail} in the five-year period between 2012 and 2016, a total of 10,339 articles and 663,173 comments from 303,665 comment threads. The corpus provides a pairing of articles and comments, together with reply structures in the comments and other metadata. The comments are those that were posted on the paper's website, already moderated through a combination of automatic moderation and flagging by other commenters.

SOCC was initially published with a small subset of labeled comments (1,043 comments), which contain constructiveness and toxicity annotations obtained through crowdsourcing. We will refer to this previously annotated subset as \smCTC (for `annotated'). 
\smCTC contains binary annotations for constructiveness (constructive or non-constructive) and a four-level toxicity classification.

Our contribution, \CTC, extends constructiveness annotations to a larger subset of 12,000 comments and introduces a refined annotation scheme that 
 captures sub-characteristics of constructiveness. The comments drawn are top-level comments (i.e., head comments and not replies) from comments threads.  
 The dataset helps fill a gap in this research area. As \citet{Vidgen19-CAF} have pointed out, not enough datasets of adequate quality exist for the task of abusive language detection. The situation is even worse for the task of constructive comment classification. One of them, the SENSEI Social Media Annotated Corpus \citep{barker-summarizing-2016} contains only 1,845 comments from 18 articles. The Yahoo News Annotated Comments Corpus (YNACC) \citep{Napoles17-FGC} is much more extensive, at 9,200 comments and 2,400 threads, capturing  characteristics such as sentiment, persuasiveness or tone of each comment. Thread-level annotations in YNACC label the quality of the overall thread such as whether the conversation is constructive and whether the conversation is positive/respectful or aggressive. While useful, this corpus does not contain constructiveness levels for each comment, but for the entire thread. C3 contributes annotations for each comment, with constructiveness labels and constructiveness sub-characteristics, as we describe below.

\subsection{Annotating constructiveness}

We used Figure Eight,\footnote{\url{https://www.figure-eight.com/}} formerly known as CrowdFlower, as our crowdsourcing interface. Contributors read the presented comment, read the article the comment refers to, identify constructive and non-constructive characteristics in the comment, and label the comment as \textit{constructive} or \textit{non-constructive}. 
Crowdsourcing was the natural choice, given the large number of comments that we wanted to annotate.

Inspired by ideas from the 
literature on comments and constructive conversations \citep{Napoles17-FGC,Niculae16-CMO,Zhang17-COD,Diakopoulos15-PTN}, from news value theory in journalism \citep{Galtung65-TSO,Weber14-DIT} and from research on civility and incivility online \citep{Coe14-OAU,Papacharissi02-TVS}, we operationalize constructiveness in terms of the presence of a number of constructive characteristics and the absence of non-constructive characteristics. In particular, our annotation scheme consists of the following attributes: 

\begin{itemize}
    \item AGREE: whether the contributor agrees with the views expressed in the comment (yes, no, partially, no opinion)
    \item CONSTRUCTIVE CHARACTERISTICS: characteristics indicating constructiveness in the comment
    \begin{itemize}
        \item provides a solution (solution)
        \item targets specific points (specific\_points)
        \item provides evidence (evidence)
        \item provides a personal story or experience (personal\_story)
        \item contributes something substantial to the conversation and encourages dialogue (dialogue)
        \item does not have any constructive characteristics (no\_con)
    \end{itemize}
    
    \item NON-CONSTRUCTIVE CHARACTERISTICS:  
    \begin{itemize}
        \item not relevant to the article (non\_relevant)
        \item does not respect the views and beliefs of others (no\_respect)
        \item is unsubstantial (unsubstantial)
        \item is sarcastic (sarcastic)
        \item is provocative (provocative)
        \item does not have any non-constructive characteristics (no\_non\_con)
    \end{itemize}
    \item CONSTRUCTIVE: overall whether the comment is constructive or not 
    \item COMMENTS: any comments or suggestions by the contributors 
\end{itemize}

We annotated the extracted 12,000 top-level comments in 12 separate batches, each batch containing 1,000 annotation units. Each unit was annotated by three to five experienced, higher accuracy contributors (referred to as \textit{Level 2 contributors} in Figure Eight terminology). 
We paid 8 cents per judgment and, as we were interested in the verdict of native speakers of English, we limited the allowed demographic region to the following four majority English-speaking countries: Canada, United States, United Kingdom and Australia. 

To maintain the annotation quality, Figure Eight uses \textit{gold questions}, which allow it to measure the performance of each contributor on the annotation task and automatically remove contributors who perform poorly. We created a pool of 300 gold questions, making sure to include gold questions with specific constructive and non-constructive characteristics. For each annotation batch we randomly chose between 90 to 130 gold questions. We set the threshold that requires the contributors to maintain the accuracy of 70\% on gold questions.  
We also included secret gold questions, which were used for our internal quality evaluation.

\subsection{Data quality}

Reliably measuring inter-annotator agreement of crowdsourced data is still an open problem, as different sets of contributors annotate different sets of questions \citep{snow-etal2008,Card18-TIO,Jagabathula17-IUA,Kiritchenko16-CRF,Mohammad-P18-1017}. 
We measure the reliability of our annotated data using four methods: examining the proportion of instances where the contributors agree; calculating chance-corrected inter-annotator agreement; evaluating crowd performance on secret gold questions; and evaluating crowd answers against expert annotations. 

First, despite the subjective nature of the phenomenon, we observed that 66.57\% instances of the total 12,000 instances had unanimous agreement among the three to five contributors on the constructiveness question and only about 10\% of the instances had serious disagreement, suggesting that humans do have common intuitions about constructiveness in news comments. 

Second, the average chance-corrected inter-annotator agreement for the binary classification (constructive or not), measured using Krippendorff's $\alpha$, for the 12 annotation batches was 0.71, suggesting that constructiveness can be annotated fairly reliably with our annotation scheme. This  number is much better than the Krippendorff's $\alpha$ of 0.49 in \citet{Kolhatkar17a}'s constructiveness corpus\footnote{\citet{Kolhatkar17a} do not report chance-corrected agreement. We calculated it for this paper.} or datasets released for other conversational attributes like toxicity \citep{Thain2017} ($\alpha = 0.59$) and personal attacks \citep{wulczyn:2016} ($\alpha = 0.45$). 

Third, we had kept aside 20 secret gold questions and measured to what extent aggregated crowd answers matched the answers of these gold questions. 
These secret questions were not labelled as gold questions in Figure Eight, and thus we were confident that the annotators could not guess that they were quality controls.
We observed that the crowd agreed with the secret gold questions 90\% to 100\% of the time. 

Finally, we examined the quality of the crowd annotations with expert evaluations. We asked a professional moderator, with experience in creating and evaluating social media content, to annotate a sample of 100 randomly selected instances. We aggregated the crowd annotations and compared them with expert answers. Overall, the expert agreed with the crowd 87\% of the time on the constructiveness question.
Among the 13\% of the cases where the expert did not agree with the crowd, the majority of the cases were marked as constructive by the crowd and non-constructive by the expert. This may be because the expert had a more critical eye, informed by her experience as a moderator. For instance, in Example \ref{ex:disagreement1}, a comment on an article about the conflict between religious accommodation in separating men from women and the right to gender equality,\footnote{\url{http://www.theglobeandmail.com/opinion/my-quarrels-not-with-york-but-ontarios-rights-code/article16350272/}} the crowd workers labelled it as constructive, perhaps because of its strong stance on equality. The expert annotator pointed out that the comment does not open dialogue. 

Example \ref{ex:disagreement2} is a response to an editorial\footnote{\url{https://www.theglobeandmail.com/opinion/editorials/why-the-premier-of-alberta-shouldnt-get-to-decide-who-is-media/article28775443/}} that criticizes the Alberta provincial government for banning a right-wing media outlet from its news conferences. The assessment of the expert annotator shows that she has read the comment and the editorial very carefully and believes that the commenter did not understand its nuances.

\begin{small}
\begin{exe} 
    \ex \label{ex:disagreement1} Comment: Any kind of segregation is abhorrent and cannot be accepted. 
    \glt Expert annotator: Short. No solution provided, nothing shared to encourage active participation. Doesn't leave room to look at views different from commenter's own. 
    
    \ex \label{ex:disagreement2} Comment: What credentials and where do they come from? who issues them? You can't possibly accommodate the whole herd of people who call themselves `media' today, thanks to the Internet. Someone needs to look at this question and come up with an answer for all governments. Sounds like the Alberta NDP plan to do just that: maybe you should all get on board and make a decision that is followed across the country.
    \glt Expert annotator: I can now see how this could have been viewed as constructive, because the questions posed by the commenter relate to the article, but at the same time, the questions included in the comment, and the comment itself  show that the commenter either didn't actually read the article or understand the points being made.
\end{exe}
\end{small}

\subsection{Corpus analysis}

We aggregated the responses of all annotators for all of the questions in our annotation scheme. For the constructiveness question, we assigned an \textit{aggregation score} in the range 0.0 $\leq$ \textit{score} $\leq$ 1.0. Then each instance is assigned a constructiveness label based on a threshold of 0.5; the instances with \textit{score} $ > 0.5$ were labeled as \textit{constructive} and others were  labeled as \textit{non-constructive}. The resulting corpus is slightly higher in constructive comments (6,516) than non-constructive comments (5,484). Among all 12,000 instance, 89.7\% instances had a clear consensus among annotators. The remaining 10.3\% (1,238) comments had aggregation scores in the range $0.4 \leq$ \textit{score} $\leq 0.6$, suggesting that there was no clear consensus among annotators on these comments. 

Since we were interested in examining how constructiveness and toxicity interact with each other, we enriched our corpus with toxicity scores provided by the Perspective system, a proprietary text scoring algorithm by Jigsaw that produces a number of attributes related to toxicity.\footnote{\url{https://www.perspectiveapi.com/}} 

\begin{table}[htb]
\centering
\begin{tabular}{llr}
\toprule
& \textbf{Characteristic}	&	\textbf{\# of Comments} \\ 
\midrule
\multirow{7}{*}{Constructive}   & constructive & 6,516 \\
& dialogue         & 7,704  \\ 
& solution         & 5,741  \\ 
& specific\_points & 6,897  \\ 
& personal\_story  & 4,217  \\ 
& evidence         & 5,551   \\ 
& con\_other & 4 \\
\midrule
\multirow{7}{*}{Non-constructive}   & non\_constructive & 5,484 \\
& provocative      & 5,557  \\
& sarcastic        & 4,685  \\
& no\_respect      & 3,043  \\
& unsubstantial    & 7,532  \\
& non\_relevant    & 3,833  \\
& noncon\_other & 2 \\
\bottomrule \\
\end{tabular}
\caption{Description of the \CTC dataset. The `constructive'/`non\_constructive' numbers are top-level binary labels where a majority of annotators found the comment constructive or not. For the sub-characteristics, we increase the counter when at least one annotator finds the attribute in the comment.}
\label{tab:data_description}
\end{table}

\subsubsection{Constructive and non-constructive characteristics}
Our annotation scheme decomposes the notion of constructiveness into a taxonomy of constructive and non-constructive sub-characteristics. 
Table \ref{tab:data_description} provides a breakdown of the distribution of these characteristics in the \CTC dataset.

\begin{figure}[htb]
    \centering
    \subfloat{{\includegraphics[scale=0.5]{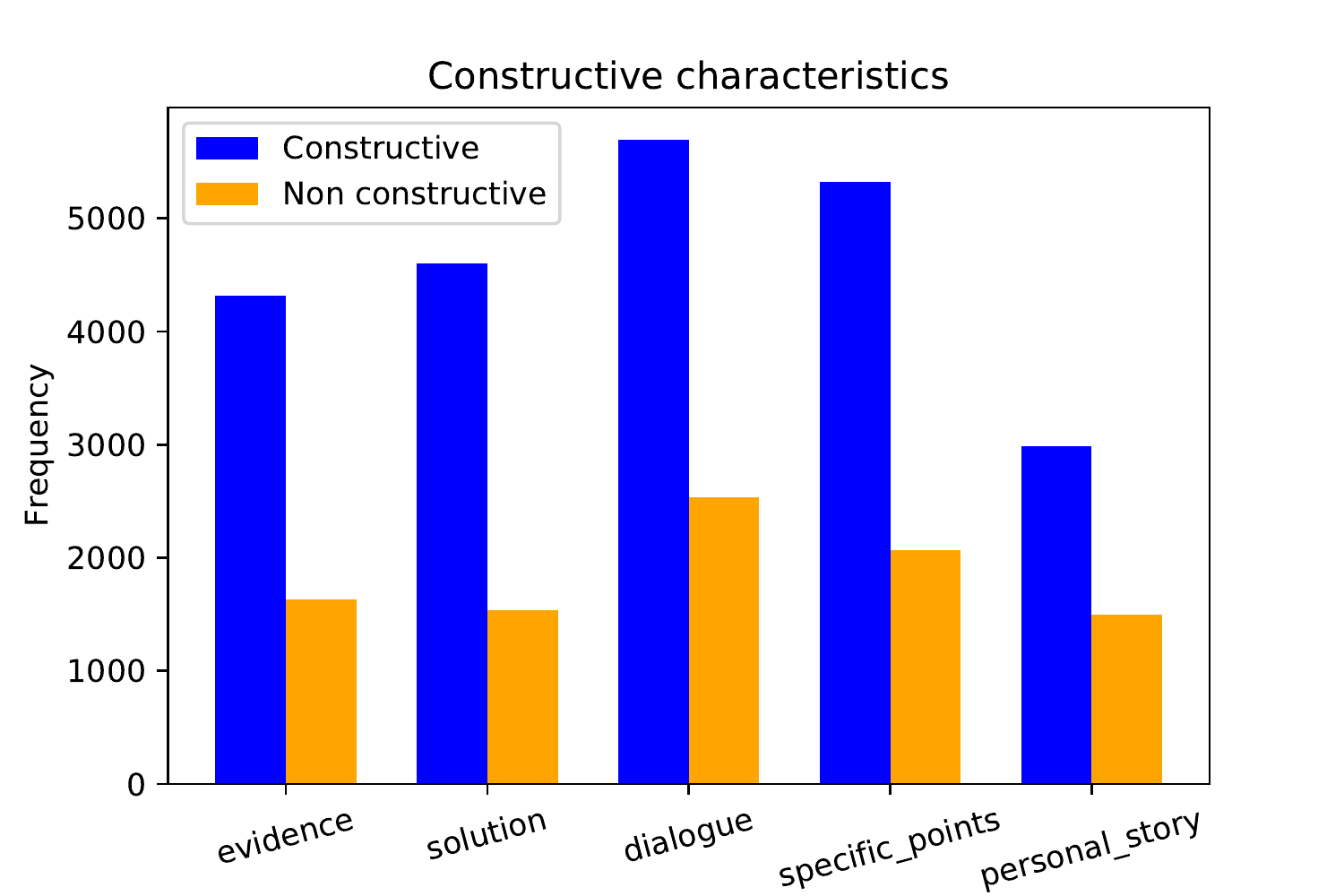} }}%
    \qquad
    \subfloat{{\includegraphics[scale=0.5]{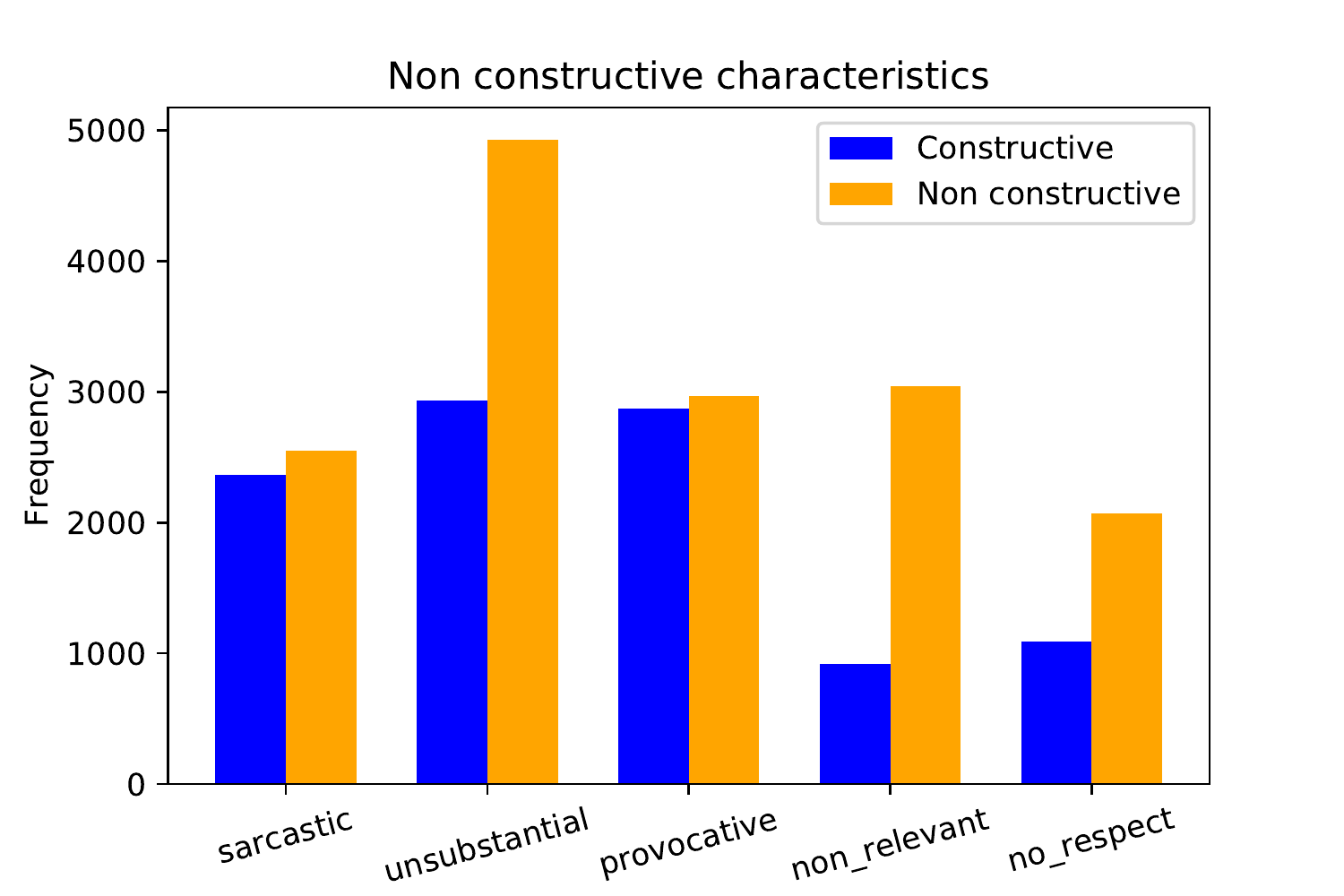} }}%
    \caption{Distribution of constructive and non-constructive characteristics in 6,516 constructive and 5,484 non-constructive comments.}%
    \label{fig:constructive_chars_dist}
\end{figure}

A question that arises is how well these characteristics perform at capturing the broader attributes of constructive and non-constructive comments. 
Figure \ref{fig:constructive_chars_dist} 
breaks down the presence of the sub-characteristics in constructive and non-constructive comments.
Constructive characteristics have a higher prevalence among constructive comments and vice versa for non-constructive characteristics. 
Indeed, there are no constructive comments that lack some constructive characteristic and likewise for non-constructive. 
While we do include a `catch-all' bucket of other characteristics, we found its use negligible. This suggests that these attributes form a comprehensive set of necessary conditions for the presence of constructiveness or non-constructiveness.\footnote{Because our survey asked raters multiple questions, it is possible that there was some effect on the raters that would not have happened if they had been asked the questions independently; investigating these potential effects is a possible branch of further work.} However, we found that they are not sufficient, as a comment can have one or more constructive sub-characteristics without being constructive.

In order to understand the impact of each of the characteristics on whether or not a comment is found to be constructive, we conduct a logistic regression, normalizing for the standard deviation of each of the characteristics. 
We are able to achieve an F1 value of 0.87 on the logistic regression, again demonstrating that the attributes provide a good breakdown of constructiveness. 
Table \ref{tab:logistic_regression} 
lists the coefficients of each of the characteristics in the logistic regression. 
The coefficients can be interpreted as the expected change in the log odds of a comment being considered constructive if one more annotator believed it had the corresponding characteristic.
From this we see that \textit{dialogue} is the most important predictor of whether a comment is found to be constructive and, for example, that dialogue has 47\% more impact on the log-odds of constructiveness than \textit{personal story}. 
Similarly, we see that a comment being \textit{provocative} does not have much of an impact on its likelihood of being found non-constructive and, indeed, being irrelevant or unsubstantial are much more significant contributors. In summary, the sub-characteristics provide a useful set of criteria for moderation, whether manual or automatic. 



\begin{table}[htb]
\centering
\begin{tabular}{lrr}
\toprule
\textbf{Variable}	&	\textbf{Coefficient} & \textbf{CI}\\
\midrule
dialogue &  0.77 & (0.71, 0.83) \\
solution &  0.74 & (0.67, 0.80) \\
specific\_points & 0.59 & (0.53, 0.65) \\
evidence & 0.58 & (0.51, 0.64) \\
personal\_story & 0.53 &  (0.46, 0.59) \\
provocative & -0.38 & (-0.44, -0.31) \\
no\_respect & -0.39 & (-0.45, -0.32) \\
sarcastic & -0.42 & (-0.48, -0.35) \\
non\_relevant & -0.84 & (-0.91, -0.78) \\
unsubstantial & -1.17 & (-1.23, -1.10) \\
\bottomrule \\
\end{tabular}

\caption{Coefficients of normalized logistic regression to predict constructiveness.}
\label{tab:logistic_regression}
\end{table}

\subsubsection{Agreeing on the views in the comment}
Intuitively, we might expect annotators to be predisposed to attach the constructive label to comments they agree with. Indeed, when examining the  correlation between constructiveness and agreement we get a Pearson correlation coefficient of 0.56 (moderate correlation). This could lead us to believe that constructiveness is really a marker for whether an annotator agreed with the comment rather than a statement about the intrinsic quality of the comment. 

In order to differentiate between these two cases, we looked at the `controversial' comments for which we had at least one annotator agree with the comment and one annotator disagree. For each of these comments, we selected a random agreeing annotator and a random disagreeing annotator. We calculate the inter-annotator agreement between these conflicting annotators on the set of controversial comments. If constructiveness was only a proxy for agreement, we would expect low inter-annotator agreement in this experiment. Instead, we found that the percentage agreement (i.e., the fraction of pairs which gave the same value for whether the comment was considered constructive) was 81.2\% (with a Krippendorff's $\alpha$ score of 0.57). Thus, we can establish that the majority of the time, even for comments with disagreement among annotators, constructiveness is measuring something qualitatively different from whether the annotator agrees with the comment.

\subsubsection{Constructiveness and toxicity}
\label{subsec:con_toxicity}

We are also interested in understanding the connection between constructiveness and toxicity. As mentioned in Section \ref{sec:related} earlier, toxicity is a negative characteristic of online conversation that has garnered significant research attention. A priori, one might expect there to be a strong negative relationship between constructiveness and toxicity, with constructive comments usually being non-toxic and vice-versa. This would mean that we could rely on existing toxicity detection systems to detect constructiveness.

\begin{figure}[htb]
    \centering
    \includegraphics[scale=0.80]{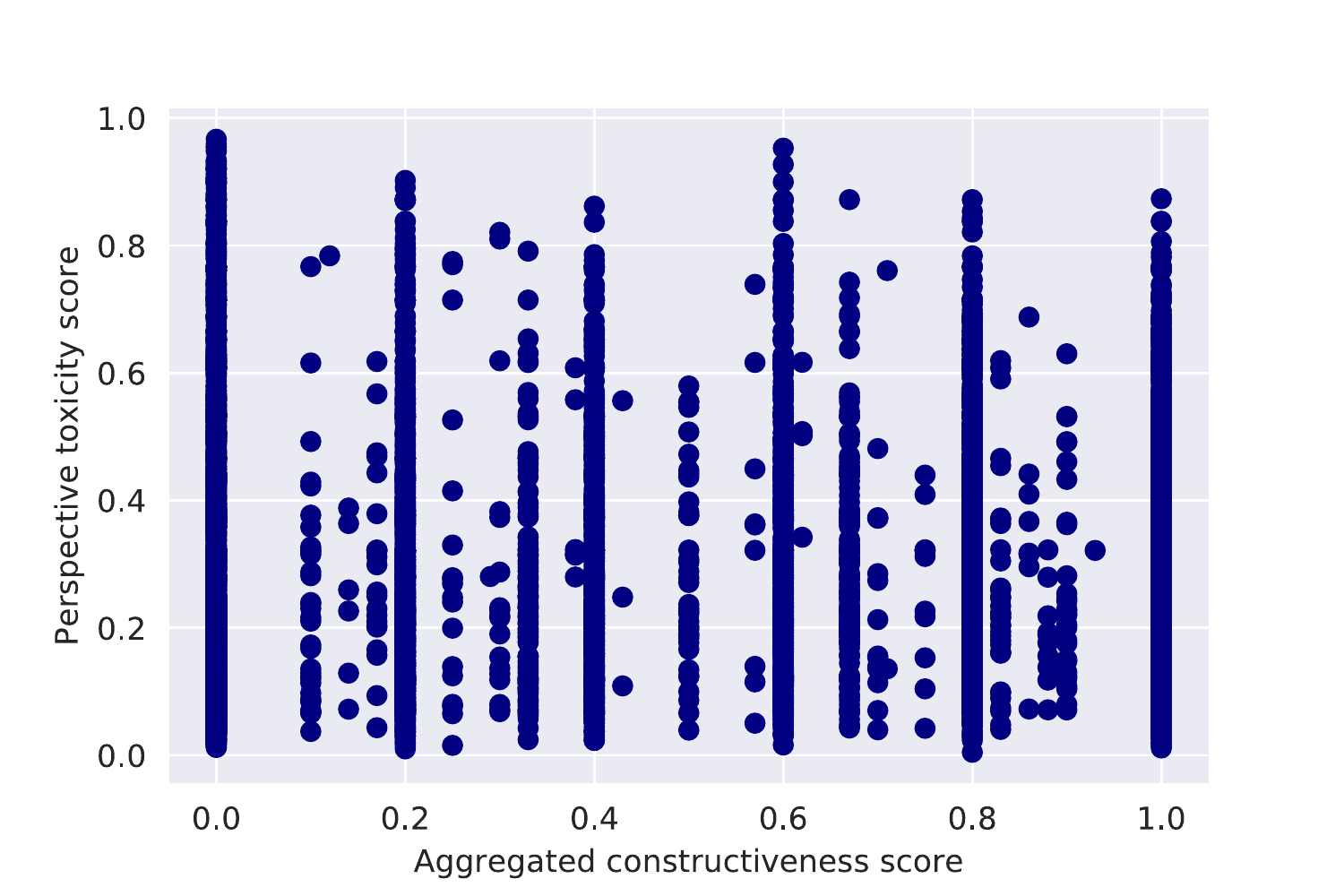}
    \caption{Constructiveness and toxicity.}
    \label{fig:toxic-constr}
\end{figure}

To understand this relationship, we looked at the correlation coefficient between the aggregated constructiveness scores and the toxicity probabilities given by the Perspective system. We calculated the correlation relationship between the variables (Pearson $= -0.02$, Spearman $= 0.04$, Kendall $\tau$ $= -0.04$), which is also demonstrated in the scatter plot shown in Figure \ref{fig:toxic-constr}. Additionally, in Section \ref{sec:experiments} we demonstrate that even a large set of toxic features including toxicity, identity based hate, insults, obscenity and threats achieve relatively low classification scores in modeling constructiveness. We conclude, then, that constructiveness and toxicity are different features. 
These results are in line with 
\citet{Kolhatkar17b}'s observation that constructiveness and toxicity are orthogonal.\footnote{Note that the comments in our dataset are moderated and they rarely contain very toxic comments.}

\section{Modeling constructiveness}
\label{sec:modeling}

In this section we describe our computational models for identifying constructiveness. We treat the problem of identifying constructive comments as a binary classification problem and investigate the characteristics of constructive comments. 
We explore classical feature-based models as well as three popular deep learning models: Long Short-Term Memory networks (biLSTMs) and Convolutional Neural Networks (CNNs) with pretrained GloVe embeddings\footnote{\url{https://nlp.stanford.edu/projects/glove/}} and pretrained Bidirectional Encoder Representations from Transformers (BERT) \citep{bert-2019}. The implementation details for these architectures can be found in \ref{sec:deep_architectures}.

For the classical feature-based models, we model constructiveness in terms of a number of automatically-extracted features.
Our features are primarily derived from two sources:  features used by \citet{Kolhatkar17b} and toxicity-related scores given by Jigsaw's Perspective system. We also include some additional linguistic and text quality features that we believe are relevant for the phenomenon. All the features are summarized in Table \ref{tab:constructiveness_feats}. With these features, we trained sklearn's linear Support Vector Machine (SVM) classifier with Stochastic Gradient Descent learning.\footnote{We also experimented with other popular classifiers such as Logistic Regression, XGBoost, and Random Forest. We chose SVMs because all of these classifiers performed comparably on our task.} We briefly describe the features used in our model as follows. 

\begin{table*}[htb]
{\footnotesize
\centering
\begin{tabular}{p{3cm} p{13cm}}
\toprule
\textbf{Feature class}	&	\textbf{Description}\\
\midrule
Lexical$^*$  (2)     &   1- to 3-gram counts and 1- to 3-gram TF-IDF weighted phrases\\
\midrule
Length$^*$ (4)	&	Length of the comment: number of tokens in the comment, number of sentences, average word length, average number of words per sentence \\
\midrule
Argumentation$^*$ (5) &	Presence of discourse connectives (\textit{therefore, due to})\\
							&	Reasoning verbs (\textit{cause, lead}), modals (\textit{may, should})\\
                            &	Abstract nouns (\textit{problem, issue, decision, reason})\\
                            & Stance adverbials (\textit{undoubtedly, paradoxically)}\\
\midrule                            
Named-entity$^*$ (1)	&	Number of named entities in the comment\\
\midrule
Text quality$^*$ (5) & Readability score, personal experience description score, number of spelling mistakes, number of CAPS words, number of punctuation tokens\\
\midrule
Content quality$^\dagger$ (3) & Text coherence score, unsubstantial probability, and spam probability\\
\midrule
Aggressiveness$^\dagger$ (3) &  Aggressiveness expressed in the comment: attack on the author, attack on fellow commenter, attack on the publisher\\
\midrule    
Toxicity$^\dagger$ (8) & Toxicity in the comment: severe toxicity, sexually explicit, toxicity, identity hate, insults, obscenity, threats, inflammatory, likely to reject\\
\bottomrule \\
\end{tabular}
\caption{Automatically-extracted constructiveness features. Features marked with * are from \citet{Kolhatkar17b}. Features marked with $\dagger$ are the scores given by Jigsaw's Perspective system.}
\label{tab:constructiveness_feats}
}
\end{table*}

\paragraph{Linguistic features}
We incorporate features from \citet{Kolhatkar17a} that capture linguistic aspects of the comment. In particular, we include lexical features, length features, features that are present in argumentative text (e.g., connectives or stance adverbials), named-entity features, and text quality features.

\paragraph{Perspective toxicity-related scores}
The Perspective system from Jigsaw is a proprietary text scoring
algorithm that produces a number of attributes. It is
trained on data similar to that published in the Kaggle Toxic
Comment Classification Challenge,\footnote{\url{https://www.kaggle.com/c/jigsaw-toxic-comment-classification-challenge}}
where many participants produced
high performance models using a variety of neural network techniques.
The Perspective system produces a number of scores and we choose 14 of these scores relevant for our task. We organize these 14 features into three groups representing different aspects of constructiveness: content-quality features, aggressiveness features and toxicity features. Content-quality features include probabilities representing text coherence, whether the comment is substantial, and the probability of it being spam. The second group has three features, all related to aggressiveness expressed in the comment: attack on the author, attack on the fellow commenter and attack on the publisher. The third group has eight toxicity related features: severe toxicity, sexually explicit, toxicity, identity hate, insults, obscenity, threats, inflammatory and likely to reject.\footnote{For more information on these features, please refer to the Perspective documentation: \url{https://github.com/conversationai/perspectiveapi}.}

\paragraph{Additional features}
 We add three new features in the text quality feature set: number of spelling mistakes, number of capitalized words, and number of punctuation tokens.

\section{Experiments} \label{sec:experiments}
We carried out four sets of experiments: 

\begin{enumerate}
    \item Benchmark experiments to compare the models trained on \CTC to those of \citet{Kolhatkar17b}.
    \item Feature set experiments to gain insight into the most important features that characterize constructiveness.
    \item Domain adaptation experiments to understand how generalizable models trained in one domain are to other domains.
    \item Length experiments to investigate to what extent the performance of our models is attributed to length-related features and how we can build more robust models for the phenomenon. 
\end{enumerate}

The experiments were conducted using four datasets: \CTC, \smCTC, \NYT, and \YNACC. We described \CTC (12,000 annotated comments) and \smCTC (1,035 annotated comments) in Section \ref{sec:data_and_annotation} above.\footnote{Although the original \smCTC contains 1,043 instances, we had to eliminate 7 instances because they did not correspond smoothly to the comment identifiers in raw SOCC. Our experiments are conducted on 1,035 instances of \smCTC.} \CTC was split into 80\% train (\CTC train) and 20\% test (\CTC test) portions.

We use two more corpora:
the New York Times Picks Corpus (\NYT) and a subset of the Yahoo News Annotated Comments Corpus (\YNACC) ~\citep{Napoles17-FGC}. The \NYT  corpus contains 15,147 comments chosen as interesting and constructive comments by \textit{The New York Times} moderators before comments were moderated automatically.\footnote{\url{https://www.nytimes.com/2017/06/13/insider/have-a-comment-leave-a-comment.html}} These comments were extracted by \citet{Kolhatkar17b} using the NYT API.\footnote{\url{https://developer.nytimes.com/}} The \YNACC  contains 15,178 comments from non-constructive comment threads, a subset of the Yahoo News Annotated Comments Corpus (YNACC),\footnote{\url{https://github.com/cnap/ynacc}} which contains thread-level constructiveness annotations for comment threads posted on Yahoo News articles. The assumption here is that a comment from a non-constructive thread is non-constructive and vice versa, although this assumption may introduce some noise in the data.

\subsection{Benchmark experiments}

The previous best performance on the \smCTC dataset was achieved by \citet{Kolhatkar17b} who trained a linear SVM on a combination of data from the \NYT and \YNACC datasets and achieved an F1-score of $0.84$. In Table \ref{tab:prev_comparison}, we benchmark the linear SVM model trained on \CTC against this previous work and show that, despite having about one third of the data used in \citet{Kolhatkar17b}, a model trained on \CTC achieves a better model performance on the \smCTC test set.

\begin{table}[htb]
    \centering
    \begin{tabular}{lrr}
    \toprule
    \textbf{Train dataset}    & \textbf{\Centering Size}    &  \textbf{F1}\\
    \midrule
    \CTC train &   9,600 &   0.87\\
    \NYT + \YNACC  & 30,325    &    0.84\\
    \bottomrule
    \end{tabular}
    \caption{Comparison with \cite{Kolhatkar17b}. The F1 column shows the F1 score on \smCTC (size = 1,035) with a linear SVM.}
    \label{tab:prev_comparison}
\end{table}

\subsection{Feature sets experiments}
The goal of these experiments was to gain insight into the most important features that characterize constructiveness. To that end, we examined how individual feature sets contribute to predicting constructiveness.  

Table  \ref{tab:features_results} shows the results of these experiments. The first results column shows F1 scores for different models when we trained on \CTC train and tested on \CTC test. Among the feature-based models, our four length features are the best predictors of constructiveness. This is not a surprise given the skewed distribution of length in constructive and non-constructive comments (see Figure \ref{fig:lengths} below).
The text-quality and all features also demonstrate comparable performance. Perspective's aggressiveness and toxicity features do not seem to be adequate predictors for constructiveness, confirming our results from Section \ref{subsec:con_toxicity} that toxicity is orthogonal to constructiveness and presence or absence of toxicity is not a strong indicator of constructiveness. 

\begin{table}[htb]
    \centering
    \begin{tabular}{llccc}
    \toprule
    \multirow{2}{*}{\textbf{Model}} & \textbf{TRAIN}   & \multicolumn{1}{c}{\textbf{\CTC train}}    & \multicolumn{1}{c}{\textbf{\CTC train}}   & \multicolumn{1}{c}{\textbf{\NYT+YNACC*}} \\
    & \textbf{TEST} & \multicolumn{1}{c}{\textbf{\CTC test}} & \multicolumn{1}{c}{\textbf{\NYT+YNACC*}}  &   \multicolumn{1}{c}{\textbf{\CTC test}}\\
    \midrule
    Lexical  &  & 0.82    & 0.81  & \textbf{0.84}\\
    Length  &  & \textbf{0.93}  & 0.82  &  0.76\\
    Argumentation  & & 0.76  & 0.69  & 0.75\\
    Named-entity & & 0.73  & 0.72  &  0.73\\    
    Text quality & & 0.90  & 0.82  &   0.81\\
    Content quality & & 0.88  & 0.79  &   0.78\\
    Aggressiveness & & 0.60  & 0.75  &   0.61\\
    Toxicity & & 0.67  & 0.66 &   0.67\\
    All features & & 0.91  & 0.82  &   0.81\\
    \midrule
    biLSTM  & & \textbf{0.93}  & 0.83  &  0.80 \\
    CNN & & 0.92 & 0.83 &  0.72 \\
    BERT & & \textbf{0.93}  & \textbf{0.84} &  0.78 \\
    \bottomrule 
    \end{tabular}
    \caption{Feature and domain adaptation results. Each cell shows the F1 score with the given model and train/test setting.}
    \label{tab:features_results}
\end{table}

\subsection{Domain adaptation experiments}
The second set of experiments was conducted to examine how transferable the learned models are on different datasets and whether the models are capturing general characteristics of constructiveness or topic-specific characteristics. \CTC contains comments posted on articles from a Canadian national newspaper, which are likely to discuss Canadian issues and politics, whereas \NYT and \YNACC contain comments posted on \textit{The New York Times} and Yahoo News articles, respectively, which are more likely to address American issues and politics. 
We examined how features learned on one dataset perform on the other. The second results column (\CTC train and \NYT + \YNACC test) and the third results column (\NYT + \YNACC train and \CTC test) in Table \ref{tab:features_results} show the outcome of these experiments. While length is the most important feature in the single domain context, when we move to a new context where the training and test sets differ, its relative importance decreases. In the domain transfer context, text quality and lexical features play an important role. 

In case of the deep models, the performance drops markedly when we move to a new context, but they are still the best performing models when trained on \CTC train and tested on \NYT + \YNACC (second results column of Table \ref{tab:features_results}). When trained on \NYT + \YNACC and tested on \CTC test, CNNs and BERT perform poorly. This is perhaps because the training data itself is noisy. Recall that \NYT contributed the positive cases (constructive) and \YNACC the negative instances. Drawing positive and negative samples from different sources is never ideal, and likely explains the modeling mismatch.

\subsection{Length experiments}

\begin{figure}[htb]
\centerline{\includegraphics[width=14cm]{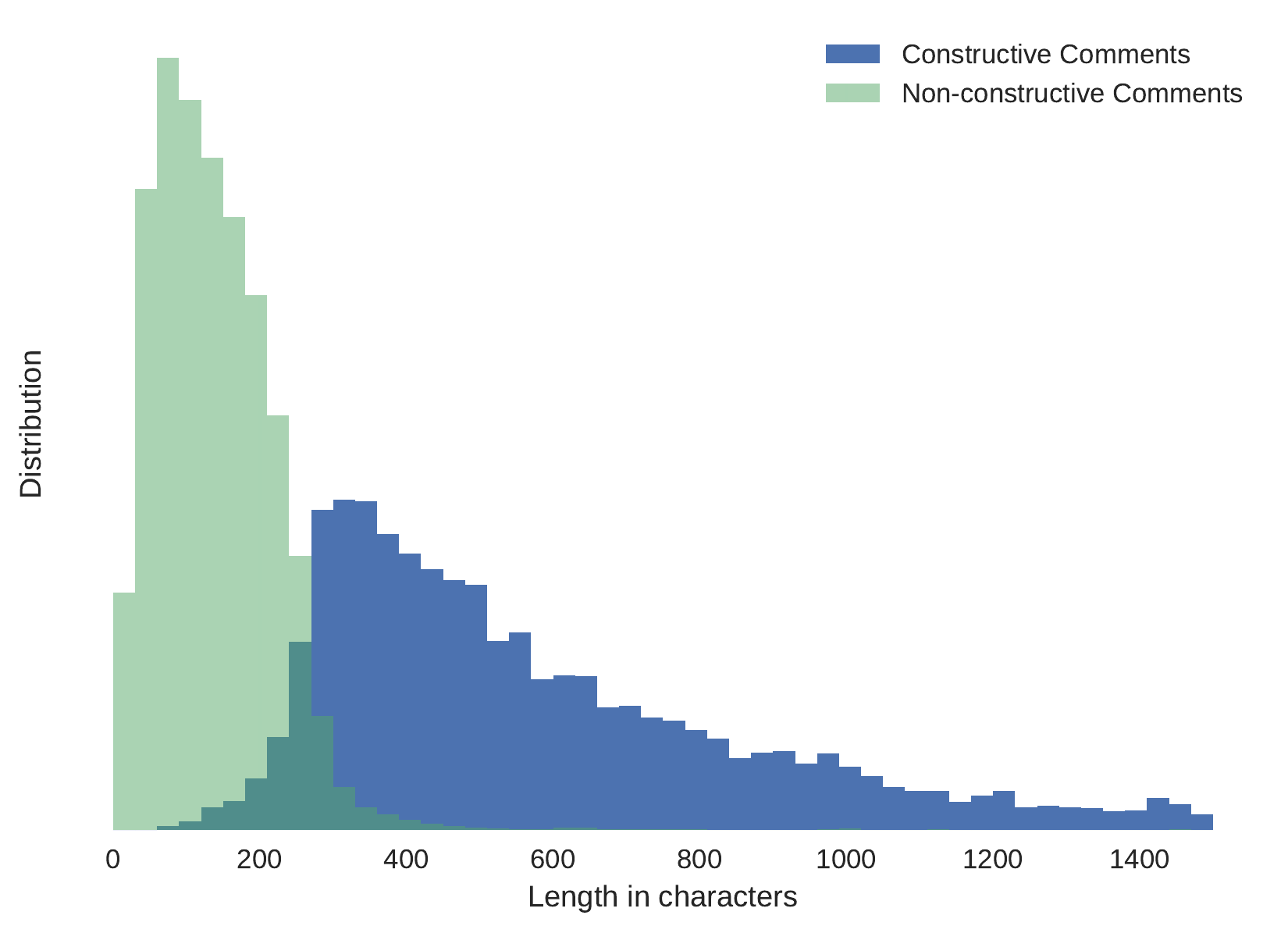}}
\caption{Distribution of comment lengths.}
\label{fig:lengths}
\end{figure}

Longer length is a characteristic feature of constructive comments; people tend to write elaborate and substantial comments when they intend to contribute something to the conversation. Figure \ref{fig:lengths} shows how this manifests for the \CTC corpus where we observed a high correlation of 0.65 between the length of the comment in words and the constructiveness label. We can observe this length correlation in the examples in Table \ref{tab:constructiveness_spectrum}. We chose those examples before we realized the importance of length for constructiveness. It seems, however, that we were intuitively inclined to correlate length and constructiveness.

It is unsurprising, then, that many of the best performing features from Table \ref{tab:features_results} have some relationship with length, though this fact has not been examined in previous work. For instance, the readability score from the text quality features is a function of length of the comment, calculated using the SMOG index formula \citep{mc_laughlin_smog_1969}, as shown in Equation (1).  
\begin{equation}
\small
1.043 \times \sqrt{\mbox{words with polysyllables} \times \frac{30}{\mbox{\# sentences}}} + 3.1291   
\end{equation}

Length, however, is not necessarily a generalizable feature for constructiveness, as seen in the domain adaptation experiments in Table \ref{tab:features_results}. It is also vulnerable to adversaries who could attempt to write long low-quality comments in order to fool the models into classifying their comments as constructive.

In this section, we evaluate our models' dependence on length as a feature. It is insufficient to simply compute the correlation of each model's predictions with comment length as we've seen that the true labels already exhibit strong correlation with length. The predictions of any well-performing model, therefore, should have some correlation with comment length. Instead, we will investigate the distribution by length of the errors that each model makes.

\begin{figure}[htb]
\centerline{\includegraphics[width=14cm]{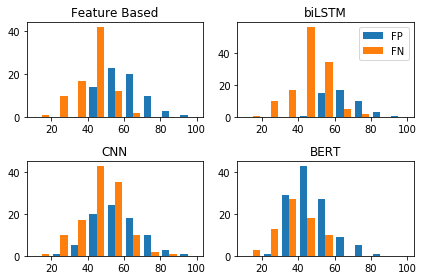}}
\caption{Histogram of model errors by length. FP = False positives. FN = False negatives.}
\label{fig:length_errors}
\end{figure}

For each model, Figure \ref{fig:length_errors} plots a histogram by comment length of the False Negative (FN) and False Positive (FP) errors exhibited by the model on the \CTC test set. It can be seen that, for every model, the false positives are distributed over comments of higher length than the false negatives, indicating an over-dependence on length as a signal. Table \ref{tab:length_experiments} confirms this insight by exhibiting that the average length of a false negative is less than that of a false positive. This gap is smallest for CNN models, demonstrating how the the max-pooling layer in this architecture reduces its ability to overfit on comment length.

 \begin{table}[htb]
    \centering
    \begin{tabular}{lcccc}
    \toprule
    \textbf{Model}  &
    \textbf{F1 on \CTC test}  &   \textbf{Length Corr.} & \textbf{FN Length} & \textbf{FP Length}\\
       &     &   (on errors) & (mean) & (mean) \\
    \midrule
     Feature based &   0.91 & 0.75 & 42 & 63  \\
     biLSTM    &   0.93 & 0.76 & 45 & 69 \\
     CNN & 0.92 & 0.32 & 49 & 59 \\
     BERT & 0.93 & 0.56 & 37 & 49 \\
    \bottomrule 
    \end{tabular}
    \caption{Results of the length experiments. Length Corr = Length Correlation. FN Length = mean comment length of false negatives. FP Length = mean comment length of false positives.}
    \label{tab:length_experiments}
\end{table}
 
 Table \ref{tab:length_experiments} also compares the performance of the models based on their F1 score and their overall dependence on length as a feature. The latter is measured by computing the correlation of the constructiveness score of each model's errors with comment length. Each model shows a positive correlation, though the effect size is dramatically smaller for CNNs than the other model types due to their architectural constraints.
 
 These experiments confirm that length insensitive deep models like CNNs are strong choices to overcome the challenge of length imbalance in constructive data. They are sufficiently flexible to benefit from being trained on the entire dataset but have inbuilt resistance to overfitting to length as a feature.

\section{Conclusions and further work}
\label{sec:conclusion}

This paper contributes: (1) a definition and operationalization of the concept of constructiveness in online comments; (2) a set of linguistic features that can be used to identify constructiveness; (3) an annotated dataset of 12,000 comments; and (4) a deep learning architecture for comment moderation.

We first conduct a thorough exploration of the problem of identifying constructive comments. This is intended to capture when a comment is seen as adding value, i.e., when it is constructive. We explored this using both crowd worker annotations and expert judgments as well as with machine learning methods. The purpose of our work is to contribute to automatic moderation, by identifying high-quality comments, rather than just filtering out undesirable comments. Constructive comments can thus be promoted or highlighted, contributing to better conversations online. 
Many news organizations try to moderate their comments not only by filtering, but also by promoting constructiveness. The New York Times, for instance, promotes constructive comments as New York Times Picks \citep{Diakopoulos15-PTN,etim-2017}.
The approach we present here, which is firmly text-based, can also be combined with metadata such as users' history, to provide a more comprehensive approach to moderation.
While our work focuses on online news comments, and on comments in response to opinion pieces, we believe that the overall constructiveness labels and the constructiveness sub-characteristics are applicable to many online conversations and commenting platforms.

The Constructive Comments Corpus (C3) is a set of 12,000 comments annotated by crowd workers, including constructive and non-constructive labels, a breakdown of which constructive and non-constructive characteristics are present in the comment, and also labels for when the human rater agreed with content of the comment.
We use this to show that, even on the subset of annotations where one rater agreed with the comment and one disagreed, the two raters agree on the constructiveness label in 81\% of the cases.
We show that, using our instructions, crowd workers can annotate constructiveness quite reliably (Krippendorff's $\alpha$ = 0.71, an improvement from 0.49 in earlier work).  

Additionally, our corpus  reveals the surprising effectiveness of text-length alone as a feature in the prediction of constructiveness.
While this feature would not produce a useful classifier---it is trivially gamed---it leads us to another fascinating result: While many machine learning models `accidentally' learn to depend on length, CNNs do not, and, moreover, they produce an effective model of constructiveness. This suggests that CNNs are a pragmatic choice of model architecture to support products that highlight constructive comments in a similar style to the New York Times' Editor's Picks.

The high quality of the \CTC annotations opens many new avenues of possible research. 
Further questions may be asked of the data. For instance, we could examine the distribution of constructiveness by topic \citep[see][for an initial exploration of topics]{Gautam19-HTC}. One could also explore whether some words or parts of speech seem to be predictive of constructive comments. Another direction is to explore the role of a conversation's context. While context is both difficult to make crowd-workers take account of, and rarely affects toxicity judgements~\citep{pavlopoulos-etal-2020-toxicity}, the relationship of constructiveness and context is still unexplored. 
With more data, other possibilities open up.
\CTC is still fairly small, especially for data-hungry deep learning models. 
Expanding this dataset would allow us to explore more complex modelling methods. 
Attention models could be used to investigate what specific aspects make a comment constructive.
The unintended biases in these models should be further probed prior to any sensitive applications.
Moreover, much additional exploration is needed to understand the key relationship between agreement with a point made within a comment and the constructiveness of the comment, to disentangle the potential effects of prior human biases.
Finally, there are likely to be other potential indicators, beyond constructiveness, of high-quality contributions in comments, such as contributing diverse points of view or healthy levels of disagreement \citep{Muddiman17-NVC,Shanahan18-JOC}. The success in modelling constructiveness suggests that there may also be significant opportunities for modelling other kinds of positive contributions to online discussion. 

The Constructive Comments Corpus (C3) is available both on Kaggle and on Simon Fraser University's online repository\footnote{\url{https://www.kaggle.com/mtaboada/c3-constructive-comments-corpus}}\footnote{\url{https://dx.doi.org/10.25314/ea49062a-5cf6-4403-9918-539e15fd7b52}} and can be cited as \citet{Kolhatkar-2020-data}. 
The code for the experiments conducted in this paper is available on GitHub.\footnote{\url{https://github.com/kvarada/constructiveness}}
We have also created an interactive demo to classify constructive comments.\footnote{\url{http://moderation.research.sfu.ca/}}

\section*{About the authors}

\noindent \textbf{Varada Kolhatkar} is a Postdoctoral Teaching and Research Fellow at the University of British Columbia. E-mail: varada.kolhatkar@gmail.com

\noindent \textbf{Nithum Thain} is a Software Engineer at Google Jigsaw. E-mail: nthain@google.com

\noindent \textbf{Jeffrey Sorensen} is a Software Engineer working on the
  Perspective API project for Jigsaw. 
E-mail: sorenj@google.com

\noindent \textbf{Lucas Dixon} is a Research Scientist at Google Research. E-mail: ldixon@google.com

\noindent \textbf{Maite Taboada} is Professor of Linguistics and Director of the Discourse Processing Lab at Simon Fraser University. E-mail: mtaboada@sfu.ca

\section*{Acknowledgements}
We thank members of the Discourse Processing Lab at Simon Fraser University for their help in testing the crowdsourcing interface and for lots of feedback. This work was supported by the Social Sciences and Humanities Research Council of Canada and by NVIDIA Corporation, with the donation of a Titan Xp GPU.

\appendix

\section{Deep Model Architectures}
\label{sec:deep_architectures}

In this appendix we outline the implementation details of the deep models discussed in Sections \ref{sec:modeling} and \ref{sec:experiments}.

Our convolutional neural network (CNN) model has a single embedding layer, a single convolution and pooling layer, and a single fully connected layer for the classification head. The embedding layers use pretrained GloVe embeddings of dimension 300 to represent the input word tokens. The convolution layer uses 128 filters each of size 3, 4, and 5 and the pooling layer performs a global max-pooling across the length of the sentence. The fully connected layer produces two values, one for each class.

The bidirectional LSTM (biLSTM) model has a single embedding layer, a single recurrent layer, and a fully connected layer for the classification head. Again, pretrained GloVe embeddings of dimension 300 are used by the embedding layer. The recurrent layer is a bidirectional LSTM with cells of size 128 for each direction. The fully connected layer produces two values.

The BERT model is built on top of the uncased variant of the pretrained BERT\textsubscript{BASE} model available on TF-Hub.\footnote{\url{https://tfhub.dev/google/bert\_uncased\_L-12\_H-768\_A-12/1}} The output sentence representation is then fed into a 3-layer fully connected neural network with layer sizes 256, 128, and 64 and a final classification head. In addition to learning the parameters of the fully connected layers, the model also tuned the top 6 layers of the BERT model, but left the remainder fixed.

At training time, all models use a dropout of 0.5 before the fully connected layers. They are trained with the adam optimizer using a learning rate of 0.001.

\section*{References}
\bibliographystyle{elsarticle-harv}
\bibliography{bibliography}

\end{document}